%% file: main.tex
\documentclass[letterpaper]{article} 
\usepackage[draft]{aaai25}
\usepackage{times}  
\usepackage{helvet}  
\usepackage{courier}  
\usepackage[hyphens]{url}  
\usepackage{graphicx} 
\usepackage{natbib}  
\usepackage{caption} 
\usepackage{algorithm}
\usepackage{algorithmic}

\usepackage{amsmath}
\usepackage{bm}
\usepackage{mathrsfs}
\usepackage{amsfonts}
\usepackage{marvosym}
\usepackage{color}
\definecolor{url_color}{RGB}{42, 83, 163}

\usepackage{newfloat}
\usepackage{listings}
\DeclareCaptionStyle{ruled}{labelfont=normalfont,labelsep=colon,strut=off} 
\floatstyle{ruled}
\newfloat{listing}{tb}{lst}{}
\floatname{listing}{Listing}
\usepackage{booktabs}
\usepackage[table,dvipsnames,svgnames, x11names]{xcolor}

\def\NickName{{GigaGS}}

\title{\NickName: Scaling up Planar-Based 3D Gaussians for Large Scene Surface Reconstruction} 

\author {
    Junyi Chen\textsuperscript{\rm 1,2,*},
    Weicai Ye\textsuperscript{\rm 1,3}\thanks{Equal Contribution.}\textsuperscript{\textrm{\Letter}},
    Yifan Wang\textsuperscript{\rm 1,2},
    Danpeng Chen\textsuperscript{\rm 3},
    Di Huang\textsuperscript{\rm 1}, \\
    Wanli Ouyang\textsuperscript{\rm 1},
    Guofeng Zhang\textsuperscript{\rm 3},
    Yu Qiao\textsuperscript{\rm 1},
    Tong He\textsuperscript{\rm 1,\textrm{\Letter}}
}
\affiliations {
    \textsuperscript{\rm 1}Shanghai AI Laboratory
    \textsuperscript{\rm 2}Shanghai Jiao Tong University
    \textsuperscript{\rm 3}State Key Lab of CAD\&CG, Zhejiang University
    \\
  \texttt{\small \urlstyle{tt}\textcolor{url_color}{\url{https://open3dvlab.github.io/GigaGS/}}}
}

\usepackage{bibentry}

\begin{document}

\maketitle

\begin{figure*}
    \begin{center}
        \includegraphics[width=\textwidth]{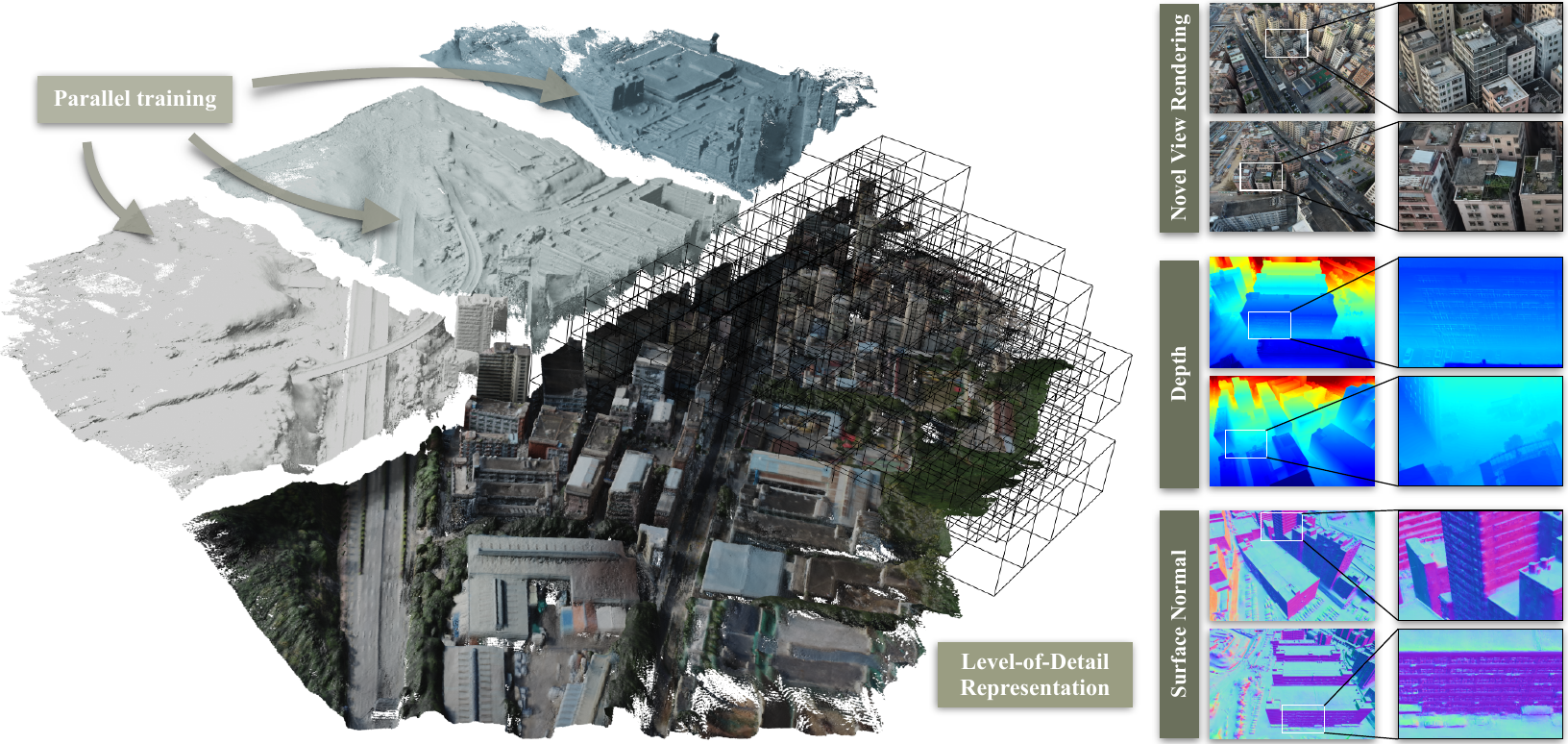}
        \caption{We propose GigaGS, the first work specifically designed for large scene surface reconstruction. 
        Our approach ensures high rendering quality while also extracting high-quality meshes.}
        \label{fig:teaser}
    \end{center}
\end{figure*}

\input{Sections/0-abstract}

\input{Sections/1-introduction}

\input{Sections/2-relatedwork}

\input{Sections/3-preliminaries}

\input{Sections/4-method}

\input{Sections/5-experiment}

\input{Sections/6-conclusion}

\bibliography{main}

\end{document}

%% file: Sections/0-abstract.tex
\begin{abstract}

3D Gaussian Splatting (3DGS) has shown promising performance in novel view synthesis. Previous methods adapt it to obtaining surfaces of either individual 3D objects or within limited scenes. In this paper, we make the first attempt to tackle the challenging task of large-scale scene surface reconstruction. This task is particularly difficult due to the high GPU memory consumption, different levels of details for geometric representation, and noticeable inconsistencies in appearance. To this end, we propose \NickName, the first work for high-quality surface reconstruction for large-scale scenes using 3DGS.
\NickName\space first applies a partitioning strategy based on the mutual visibility of spatial regions, which effectively grouping cameras for parallel processing. To enhance the quality of the surface, we also propose novel multi-view photometric and geometric consistency constraints based on Level-of-Detail representation. In doing so, our method can reconstruct detailed surface structures. Comprehensive experiments are conducted on various datasets. The consistent improvement demonstrates the superiority of \NickName. 

\end{abstract}

%% file: Sections/1-introduction.tex
\section{Introduction}
\label{sec:introduction}

3D Gaussian Splatting~\cite{kerbl20233dgs} has demonstrated remarkable performance on the task of novel view synthesis. Recently, some methods~\cite{guedon2023sugar, huang20242dgs} adapted it to surface reconstruction, which is drawing increasing attention for its numerous promising applications, such as 3D asset generation~\cite{tang2023dreamgaussian, chen2024gsgen, he2024gvgen, xu2024grm, tang2024lgm} and virtual reality~\cite{wu20234dgaussians, charatan2024pixelsplat}. 
Compared with the task of novel view synthesis, recovering the inherited 3D surfaces is much more challenging as it requires preserving 3D coherence throughout varying perspectives with only 2D projects for supervision.

Although significant advances~\cite{huang20242dgs, guedon2023sugar, yu2024gaussianopacity} have been made, these methods either focus on object-level reconstruction or struggle to capture intricate geometric surfaces.
These challenges become even more critical in the context of large-scale scene reconstruction. Firstly, the computational resource consumption is enormous, as a scene covering several square kilometers often contains billions of Gaussian points. 

Directly applying previous methods may lead to sub-optimal reconstruction quality or run into memory-related issues.
Secondly, previous neural reconstruction methods with 3D Gaussian Splatting~\cite{kerbl20233dgs} often address the challenging task by introducing monocular regularization. For example, SuGaR~\cite{guedon2023sugar} introduces single-view depth and normal geometry consistency constraints to ensure the correctness of single-view geometry.
Despite achieving good reconstruction results, these methods fail to incorporate multi-view constraints to ensure global geometry consistency. Recent work of PGSR~\cite{chen2024pgsr} first utilizes multiview geometric constraints to 3DGS representation. Although impressive, the method fails to capture the
geometric details at different scales.

To address the above challenges, we propose \NickName. To the best of our knowledge, it is the first work of high-quality surface reconstruction for large-scale scenes using 3DGS. Firstly, we implement an efficient and scalable partitioning strategy to address the computational demands of processing large-scale scenes. Unlike conventional approaches relying on spatial distance metrics, we introduce a novel grouping mechanism based on the mutual visibility of spatial regions captured by the scene cameras. This enables us to partition the scene into overlapping blocks that can be processed in parallel. 
Each block undergoes independent optimization, thus allowing for distributed processing of the scene data. Subsequently, the optimized blocks are seamlessly merged to reconstruct the complete scene, ensuring computational efficiency without compromising on reconstruction accuracy.
Secondly, we present a novel method to harness multi-view photometric and geometric consistency constraints within a Level-of-Detail (LoD) framework. This approach is designed to enhance the preservation of geometric details across different scales of the reconstructed scene. By integrating LoD representation into the constraint formulation, we ensure that the reconstruction process maintains fidelity and coherence across varying levels of scene complexity. Leveraging both photometric and geometric information from multiple views, our method facilitates robust reconstruction of intricate scene details while mitigating artifacts and inconsistencies.

To summarize, the contributions of the paper are listed as follows:
\begin{itemize}
    \item To the best of our knowledge, we are the first to utilize 3DGS for large-scale surface reconstruction.
    \item Based on a large-scene partitioning strategy, we present a novel method to add multi-view photometric and geometric consistency constraints within a Level-of-Detail (LoD) framework.
    \item Comprehensive experiments on various datasets demonstrate the effectiveness of the proposed method for large scene surface reconstruction.
\end{itemize}

%% file: Sections/2-relatedwork.tex
\section{Related Work}
\label{sec:relatedwork}

\subsection{Neural Rendering}

Neural radiance field~\cite{mildenhall2021nerf,Ye2023IntrinsicNeRF, huang2024nerf, ming2022idf} models a 3D scene by learning a continuous volumetric scene function that maps 3D coordinates and viewing directions to the corresponding RGB color and volume density. 
This approach enables the synthesis of novel views of complex scenes from a set of input images. 
Despite the numerous advancements made in recent studies \cite{fridovich2022plenoxels, chen2022tensorf, sun2022direct, mueller2022instantngp} aimed at enhancing its performance, such as reducing training duration and expediting rendering processes, the existing framework remains constrained by the absence of a clear and explicit representation. Consequently, this limitation poses significant challenges in extending its applicability to a broader spectrum of scenarios.
3D Gaussian Splatting \cite{kerbl20233dgs} is another innovative approach in neural rendering that uses Gaussian functions to represent volumetric data. 
This technique involves placing 3D Gaussians in the scene to approximate the spatial distribution of radiance and density. 
More importantly, 3DGS possesses an explicit representation that empowers real-time rendering by utilizing rasterized rendering methods. Subsequent works focus on enhancing rendering quality\cite{yu2023mipsplatting, lu2023scaffold}, further streamlining 3DGS to improve rendering speed\cite{fan2023lightgaussian}, and extending its application to reflective surfaces\cite{jiang2023gaussianshader}.

\subsection{Surface Reconstruction}

Surface reconstruction~\cite{chen2024pgsr, Ye2024FedSurfGS, Ye2024DATAP-SfM, tang2024ndsdf, ye2022deflowslam, ye2023pvo, liu2021coxgraph, li2020saliency} aims at generating accurate and detailed 3D models from various forms of input data, which is fundamental for numerous applications, including 3D modeling, virtual reality, and robotics. 
Recent advancements in deep learning have significantly improved the quality and efficiency of surface reconstruction techniques.
The utilization of neural implicit representation offers the advantage of continuous and differentiable surfaces, thereby enabling more precise and flexible reconstruction. Recent research \cite{li2023neuralangelo, guo2022neuralmanhattan, wang2021neus, yu2022monosdf} has extensively employed these representations, demonstrating their ability to generate detailed and accurate surface reconstructions capable of capturing intricate geometric details and complex topological structures.
SuGaR \cite{guedon2023sugar} successfully aligned 3DGS with the surface of the scene, yielding remarkable reconstruction outcomes. Additionally, 2DGS \cite{huang20242dgs} simplifies 3DGS, thereby facilitating a more expressive representation of the scene's structure. However, there remains scope for further enhancement in achieving quantitative results.

\subsection{large scale reconstruction}

Large scale reconstruction involves creating detailed and accurate 3D models of extensive environments, and is challenging due to the vast amount of data, the need for high precision, and the complexity of the scenes.
MegaNeRF \cite{turki2022meganerf} represents a pioneering approach for reconstructing expansive outdoor scenes. It extends the NeRF framework by partitioning the scene into manageable blocks and independently optimizing each block, thus enabling effective handling of large-scale environments. Furthermore, VastGaussian \cite{lin2024vastgaussian} proposed a blocking strategy specifically tailored for 3DGS, enabling parallel training of distinct blocks. This strategy not only reduces training time but also facilitates the attainment of high-quality rendering for the entire scene.

%% file: Sections/3-preliminaries.tex
\section{Preliminaries}
\label{sec:preliminaries}

In this work, we employ 3DGS \cite{kerbl20233dgs} as the fundamental 3D representation and rendering entity, while employing unbiased depth rendering to acquire depth maps and surface normal maps. Within this section, we shall elucidate the significance of these two technologies as essential contextual foundations.

\subsection{3D Gaussian Splatting}
\label{sec:3dgaussiansplatting}

One of the core parts in the 3DGS is the 3D Gaussian kernel, which encapsulates the visual characteristics of a spatial region. 
Each 3D Gaussian possesses several key attributes, including positional coordinates, a covariance matrix that describes the arrangement of the kernel, opacity, and spherical harmonic coefficients that encode the view-dependent colors.
During the rendering procedure, the 3D Gaussians are projected onto a 2D Gaussian distribution specific to the given viewpoint. 
Subsequently, the final rendering output for that viewpoint is generated through $\alpha$-blending:

\begin{equation}
\bm{C} = \sum_{i\in M} \bm{c}_i \alpha_i T_i, \quad T_i = \prod_{j=1}^{i-1} (1-\alpha_j),
\end{equation}

The parameters are updated via a differentiable rendering process.  

\subsection{Unbiased Depth Rendering}
\label{sec:unbiaseddepthrendering}

2DGS~\cite{huang20242dgs} represents 3D shape with 2D Gaussian
primitives. SuGaR~\cite{guedon2023sugar} adds a regularization term that encourages the Gaussians to align with the surface of the scene. It inherently provides accurate estimations of the surface normal, which corresponds to the shortest axis of the Gaussian kernel.
Inspired by these methods, PGSR~\cite{chen2024pgsr} was initially developed for the purpose of viewpoint-dependent normal vector rendering:

\begin{equation}
\bm{N} = \sum_{i\in M} \bm{R}_c \bm{n}_i \alpha_i T_i,
\end{equation}

where $\bm{R}_c$ is the rotation matrix from camera coordinates to world coordinates and $\bm{n}_i$ is the normal vector of i-th 3DGS.
Unlike previous approaches, PGSR takes a different approach by not directly rendering based on the spatial position of the 3DGS kernels. 
Instead, it assumes that the 3D Gaussian kernels can be flattened into a plane and fitted onto the actual surface. 
It then proceeds to render the distance from the camera origin to this Gaussian plane, denoted as $\mathscr{D}$:

\begin{equation}
\mathscr{D} = \sum_{i\in M} d_i \alpha_i T_i, 
\end{equation}

where, $d_i = (\bm{R}_c^T (\mu_i - \bm{T}_c)) \bm{R}_c^T \bm{n}_i^T $ represent the distance from the camera origin to $i$-th Gaussian Kernel. 
Once the distances and normals of the planes are obtained, PGSR determine the corresponding depth map by intersecting rays with these planes. 
This intersection operation ensures that the depth shapes align with the planes assumed by the Gaussian kernel, resulting in a depth map that accurately reflects the actual surfaces:

\begin{equation}
\bm{D}(\bm{p}) = \frac{\mathscr{D}}{\bm{N}(\bm{p}) \bm{K} ^ {-1} \widetilde{\bm{p}}},
\end{equation}

where $\bm{p}$ is the 2D position in the image plane. $\widetilde{\bm{p}}$ denotes the uniform coordinates of $\bm{p}$ and $\bm{K}$ is the intrinsic coordinates of the camera.

%% file: Sections/4-method.tex
\section{Method}
\label{sec:method}

The primary challenges in large scene surface reconstruction tasks are the vast area of the scene, an excessive number of fine details, and the drastic fluctuation in image brightness caused by changes in lighting and exposure factors.
Existing surface reconstruction methods \cite{li2023neuralangelo, wang2021neus, guedon2023sugar, guo2022neuralmanhattan} primarily focus on small-scale and object-centric scenes and lack explicit designs to address the challenges posed by scaling up, resulting in limited applicability to large-scale scenes. 
Some works \cite{lin2024vastgaussian, turki2022meganerf, li2024nerfxl} design complicated strategies for data partitioning, aiming to reduce training time and memory burden. However, these works mainly concentrate on the image rendering quality while disregarding the scene surface. 

To address the scalability issue in surface reconstruction tasks, we present an efficient and scalable scene partitioning strategy for parallel training of different partitions across multiple GPUs.
To capture fine-grained details across multiple levels of granularity, our framework employs a hierarchical plane representation to store different levels of details (LoD) and achieve high-quality surface reconstruction.  

Generally speaking, in Section~\ref{sec:3drep}, we present our hierarchical plane representation. 
Subsequently, in Section \ref{sec:partitioning}, we elaborate on our scalable partitioning strategy, which is based on our representation and circumvents the limitations imposed by hardware and training time, allowing us to utilize a larger number of 3D Gaussian to represent the large scene, even reaching the giga-level scale. 
Furthermore, we detail how to accurately fit the surface of the scenes in section ~\ref{sec:appearance_geometry_reg} and subsequently extract the mesh in section ~\ref{sec:meshextraction}. 

\subsection{Hierachical Plane Representation}\label{sec:3drep}

In typical 3D reconstruction tasks, training data often includes information about the same object at different scales, especially in aerial images. 
Existing works \cite{guedon2023sugar, wang2021neus, li2023neuralangelo, chen2024pgsr, yu2024gsdf, yariv2021volumesdf} struggle to directly capture features at different scales because they lack explicitly designed structures to capture levels of details.
Therefore, we introduce a new representation combining a hierarchical structure to model the surface of the scene and a flatten form closing to a planar surface.

\paragraph{Hierachical Scene Structure} 
\label{sec:hierachicalscenestructure}

\begin{figure}[h]
    \begin{center}
        \centerline{\includegraphics[width=\linewidth]{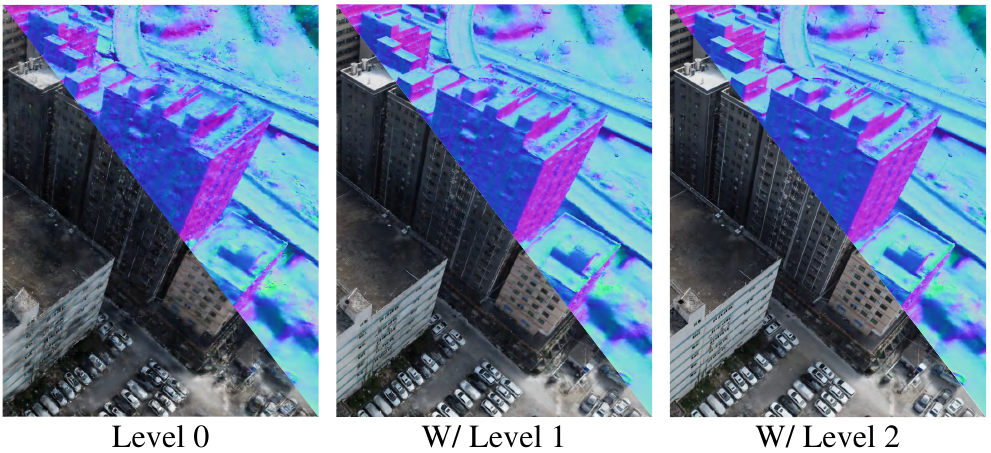}}
        \vspace{-0.2cm}
        \caption{\textbf{Visualization of the effects at different levels.} 
        We visualized the rendered images and normal maps obtained by rendering different levels of the same scene as rendering entities.}
        \label{fig:lod}
    \end{center}
\end{figure}

Given the inherent difficulties in achieving real-time rendering across various scales, especially for conventional 3DGS methods, the surface reconstruction of large-scale scenes has become a challenging task. 
We adopt a hierarchical structure, inspired by OctreeGS \cite{ren2024octreegs, lu2023scaffold}, to represent the surface of the scene, 
where a local set of $n$ 3D Gaussians is represented by a single anchor Gaussian, and during forward inference, the Multi-Layer Perceptron (MLP) is used to recover the parameters of these $n$ 3D Gaussians. The parameters of the MLP are trained jointly with the features of the anchor Gaussians. Different levels of anchor Gaussians are employed to represent features at various levels of granularity. Prior to training, it is feasible to construct a collection of anchor Gaussians at distinct levels from the point cloud $\mathbb{P}$ obtained through Structure-from-Motion (SFM) \cite{schonberger2016structurefrommotion}: 

\begin{equation}
{\rm level}_{i} = \Big\{ v_i \left\lceil \frac{\bm{p}}{v_i} \right\rfloor \Big | ~ \bm{p} \in \mathbb{P} \Big\}, ~~ 0 \leq i \leq K-1, 
\end{equation}

where $v_0$ is the fundamental voxel size, and $v_i = v_0 / k^i$ is the voxel size of level $i$ and $k$ is the fork number. 
When $k=2$, the hierarchical strategy yields the octree structure. $K$ is the maximum number of levels. 
During the rendering stage, the visibility of levels is determined based on their positional distance $d$ from the viewpoint:  

\begin{equation}
{\rm upper\_level}(d) = \max( \left\lceil \log d_{max} - \log d  \right\rfloor , K-1).
\label{eq:upper}
\end{equation}

The parameter $d_{max}$ represents the maximum distance between points in the point cloud $\mathbb{P}$ and can be calculated prior to training, and $\left\lceil \cdot \right\rfloor$ is used to denote the rounding operation.
With the distance $d$ increases, fewer 3D Gaussians with lower level are involved in the rendering process.
During the training process, we follow the approach of OctreeGS \cite{ren2024octreegs} for the addition and removal operations of anchor Gaussians.

\paragraph{Flattened 3D Gaussians} 
\label{sec:flattened3dgs} 

As the shortest axis of 3D Gaussian kernel inherently provides accurate estimations of the normal vectors \cite{guedon2023sugar},  we make efforts to compress the minimum axis of each Gaussian kernel during the training phase:

\begin{equation}
\mathcal{L}_{flatten} = \frac{1}{|\mathbb{M}|} \sum_{i\in \mathbb{M}} \big| \min(\bm{s}_i) \big|, 
\end{equation}

where $\mathbb{M}$ is the set containing the 3D Gaussians.
By doing so, we aim to constrain the shortest axis of the 3d Gaussian kernel to be perpendicular to the surface of the scene,  thereby facilitating the use of 3DGS to fit the surface of the scene. 

\begin{figure*}[h]
    \begin{center}
        \centerline{\includegraphics[width=\linewidth]{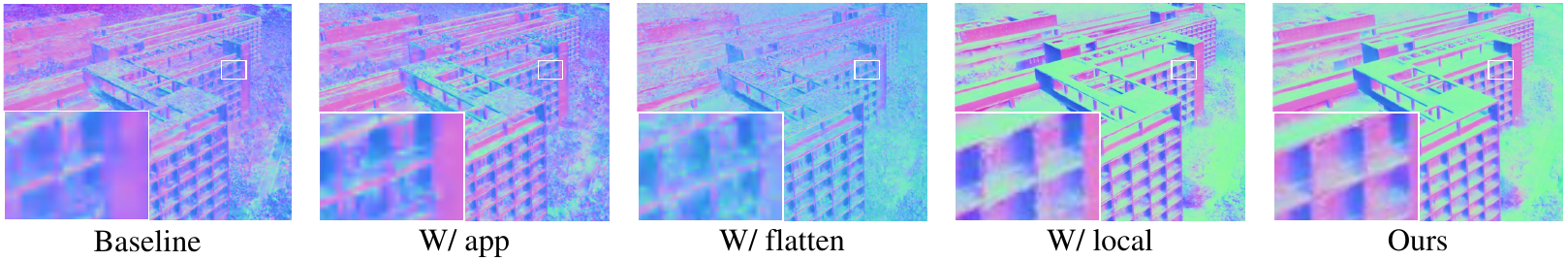}}
        \vspace{-0.1cm}
        \caption{\textbf{Different loss terms on the final optimization process.} Initially, using only the image loss as supervision does not yield a satisfactory surface reconstruction. 
        The incorporation of an appearance model reduces certain artifacts. 
        However, the addition of flatten regularization on the geometric structure without additional geometric supervision leads to a decrease in expressiveness by the model. 
        Nevertheless, the inclusion of the local loss allows for improved surface quality. 
        Finally, with the introduction of multi-view regularization, the surface reconstruction performance is further enhanced, highlighting the superiority of our method in surface reconstruction.}
        \label{fig:ablation}
    \end{center}
    \vspace{-0.8cm}
\end{figure*}

\subsection{Partitioning Strategy}
\label{sec:partitioning}

To address the challenges of scaling 3DGS to large scenes, VastGaussian~\cite{lin2024vastgaussian} proposed a partitioning strategy to evenly distribute the training workload across multiple GPUs. 
These strategies aim to overcome the limitations of 3DGS in handling large-scale scenes.
However, it still face difficulties in extreme scenarios where supervision of a specific region exists in other partitions but is not included in the training data of the current partition due to threshold-based selection strategies.
This is likely to occur because aerial trajectories change with the scene, leading to the aforementioned situation.

To tackle this issue, we propose a more robust partitioning approach that leverages the octree-based scene representation. 
Firstly, we ensure an approximately equal number of cameras in each partition $c$ by following the filtering rule of uniform camera density~\cite{lin2024vastgaussian}. Additionally, each partition is non-overlapping.
We have eliminated the manual threshold setting in visibility-based camera selection, which was proposed in VastGaussian 
Instead, we utilize the painter algorithm \cite{10.1145/800193.569954_painter} to select the cameras based on the partition anchors that can successfully project onto the camera's image plane. 
This approach aims to cover as many cameras as possible during the selection process.
To ensure sufficient supervision for each partition, we project the anchors of the partition onto the image planes of all cameras and consider the camera that can observe the anchors of the partition to be included in the training of that partition. 
This process is a greedy one without manually setting any thresholds, but it guarantees maximum supervision for each partition.

Finally, we need to expand the partitions so that each camera can render a complete image. 
Therefore, we project all anchors onto the image plane of each camera and add all visible anchors to the training set of that partition based on the equation \ref{eq:upper}.
Since we only require the anchors within the partitions for our final purpose, the expanded anchors are used solely to assist in training. 
Therefore, in Equation ~\ref{eq:upper}, we employ a floor operation rather than a round operation to reduce the number of anchors outside the partitions. 
This operation does not affect the final quality because those anchors will be discarded after training is completed.


\subsection{Appearance and Geometry Regularization}
\label{sec:appearance_geometry_reg}

As discussed in NeRF-W \cite{martinbrualla2021nerfinthewild}, directly applying existing representations to a collection of outdoor photographs can lead to inaccurate reconstruction due to factors such as exposure and lighting conditions. 
These reconstructions exhibit severe ghosting, excessive smoothness, and further artifacts.
Therefore, we introduce an appearance model to capture the variations in appearance for each image. 
Similarly, this model is learned jointly with our planar representation.
Next, to ensure that the flattened 3D Gaussians adhere to the actual surface, we enforce consistency between the unbiased depth maps and normal maps from each viewpoint. 
Simultaneously, we discover that explicit control of consistency across viewpoints has a positive impact on the final surface reconstruction quality.

\paragraph{Appearance Modeling} 
\label{sec:appearance_modeling}

We learn to model the appearance variations in each image in a low-dimensional latent space \cite{martinbrualla2021nerfinthewild}, such as exposure, lighting, weather, and post-processing effects.
In this approach, we allocate an embedding $emb_v$ for each training perspective $v$ and additionally train an appearance model $\phi$ that maps $e$ to per-pixel color adjustment values for the image.
By multiplying these adjustment values with the rendered image $\bm{I}$, we obtain a simulated lighting image that accurately represents the lighting conditions for that particular perspective:

\begin{equation}
\bm{I}_a = \phi(\bm{I}, emb_v) \bm{I}.
\end{equation}

This enables us to effectively account for the variations in lighting and appearance across different viewpoints, and the following loss is utilized:

\begin{equation}
\mathcal{L}_{app} = L_1(\bm{I}_a, \bm{I}_0) + \lambda SSIM(\bm{I}, \bm{I}_0),
\end{equation}

where $\bm{I}_0$ is the ground-truth image. $\lambda$ is utilized to adjust the relative weights between the two components. Throughout all our experiments, the value of $\lambda$ is consistently maintained at $0.25$.

\paragraph{Geometry Consistency} 
\label{sec:geo}
The vanilla 3DGS \cite{kerbl20233dgs}, which primarily relies on image reconstruction loss, tends to encounter challenges in local overfitting optimization. 
In the absence of effective regularization techniques, the intrinsic capacity of 3DGS to capture visual appearance gives rise to inherent ambiguity in the relationship between three-dimensional shape and brightness. 
Consequently, this ambiguity permits the acceptance of degenerate solutions, resulting in a discrepancy between Gaussian shape estimation and the actual surface representation \cite{zhang2020nerf++}. 
To address this issue, we have introduced a straightforward regularization constraint, aimed at enforcing geometric consistency between local depth map and surface normal map:
\begin{equation}
\mathcal{L}_{local} = \frac{1}{|\bm{I}|} \sum_{i \in \bm{I}} \Big| \frac{(P_{i,0} - P_{i,1}) \times (P_{i,2} - P_{i,3})}{|(P_{i,0} - P_{i,1}) \times (P_{i,2} - P_{i,3})|} - \bm{n}_i \Big| ~ \omega_{i} ,
\end{equation}

where $P_{i,j}$ is the 3D position of adjacent pixel $j$ at the top, bottom, left, and right positions relative to pixel $i$ in the camera coordinate system. Additionally, $n_i$ is the normal value of pixel $i$. This assumption holds notable significance as it asserts the interconnectedness of adjacent pixels within a common plane. 
For pixels that belong to depth discontinuities, we introduce an uncertainty factor, denoted by $\omega_{i}$, which serves to quantify the likelihood that the pixel $i$ belongs to the boundary of the surface:

\begin{equation}
\omega_{i} = \Big | (P_{i,0} - P_{i,1}) (P_{i,2} - P_{i,3}) \Big |.
\end{equation}

Evidently, the dot product mentioned above will yield diminished values in regions characterized by substantial depth disparities, thereby identifying them as edge regions. 
Consequently, these areas should experience a reduced influence from the constraint imposed by the depth and normal consistency.

\begin{figure*}[htbp]
    \begin{center}
        \centerline{\includegraphics[width=0.9\linewidth]{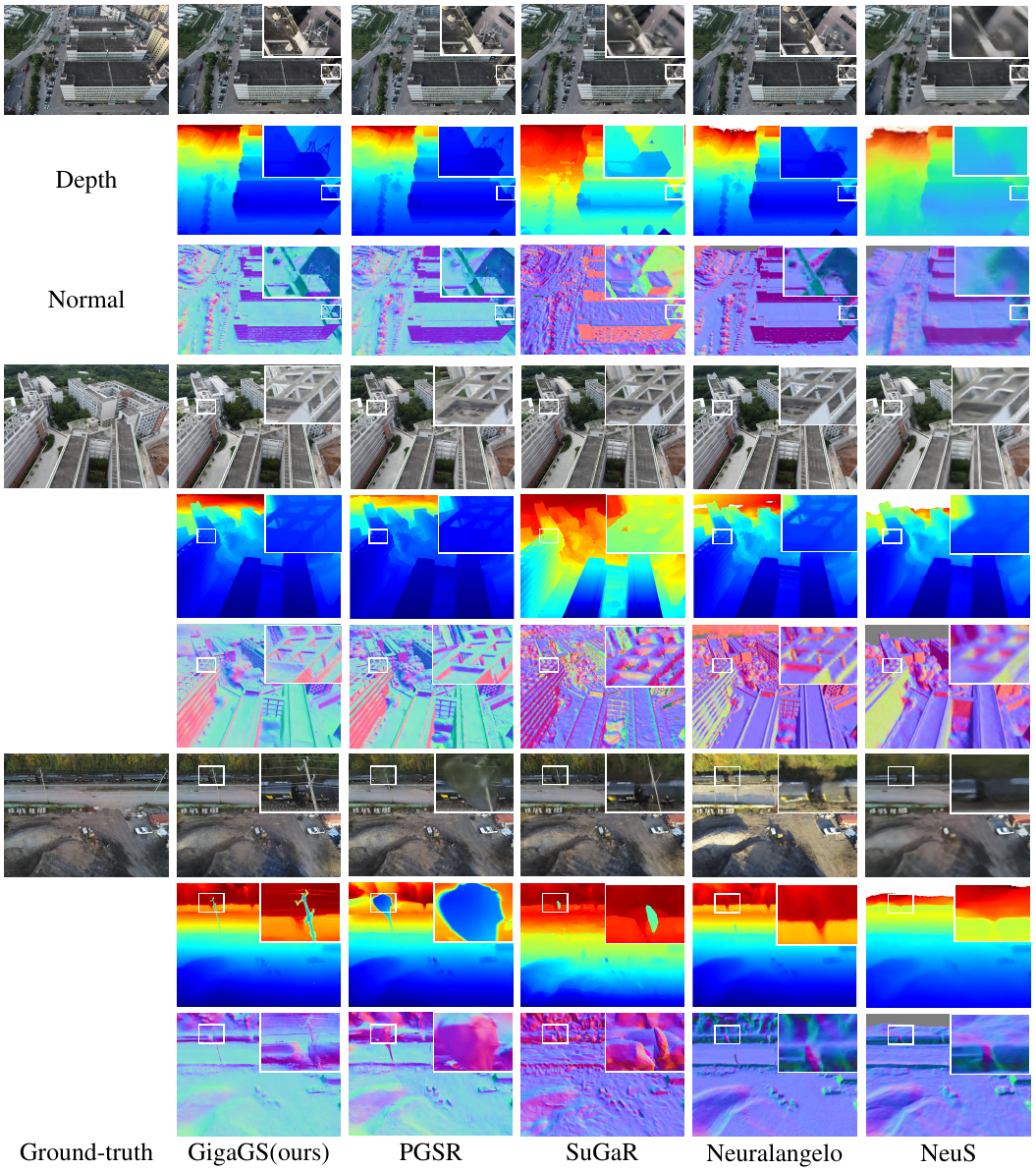}}
        \vspace{-0.1cm}
        \caption{\textbf{Comparison of visualization results.} 
        We presented rendered views of the test scenes, along with the corresponding depth maps and normal maps from the same viewpoint. 
        }
        \vspace{-0.2cm}
        \label{fig:visual}
    \end{center}
    \vspace{-0.6cm}
\end{figure*}

\paragraph{Multi-View Consistency} 
\label{sec:multiview}

Single-view geometry regularization can maintain consistency between depth and normal geometry, but the geometric structures across multiple views are not entirely consistent as shown in figure ~\ref{fig:ablation}. Therefore, it is necessary to introduce multi-view geometry regularization to ensure global consistency of the geometric structure.
We employ a photometric multi-view consistency constraint based on planar patches to supervise the geometric structure. 
Specifically, we can render the normal and the distance from each pixel to the plane. 
Then, the optimization of these geometric parameters can be achieved through patch-based inter-view geometry consistency.
For each pixel point $\bm{p}_r$, we can warp it to an adjacent viewpoint:

\begin{equation}
\tilde{\bm{p}}_n =  \bm{H}_{rn} \tilde{\bm{p}}_r, 
\end{equation}

where $\tilde{\bm{p}}$ is the homogeneous coordinate of pixel point $\bm{p}$, and homography $\bm{H}_{rn}$ can be computed as:

\begin{equation}
\bm{H}_{rn} = \bm{K}_r (\bm{R}_{rn} - \frac{\bm{T}_{rn} \bm{n}_r^T}{d_r}) \bm{K}_r^{-1}, 
\end{equation}

where $\bm{R}_{rn}$ and $\bm{T}_{rn}$ are the relative transformation from the reference frame to the neighboring frame. 
With a focus on geometric details, we convert the color image $\bm{I}$ to grayscale image $I$ to supervise our geometric parameters.
Then, we utilize the normalized cross correlation (NCC) ~\cite{yoo2009fast} of patches in the reference frame and the neighboring frame as a metric to evaluate the photometric consistency. 

\begin{equation}
\mathcal{L}_{ncc} = \frac{1}{\mathbb{|V|}} \sum_{\bm{p}_r \in \mathbb{V}} \Big(1-NCC \big( I(\bm{p}_r), I(\bm{H}_{nr} \bm{p}_n) \big) \Big).
\end{equation}

Where $\mathbb{V}$ is the valid region checked through geometric consistency constraints. 
The warp operation may introduce inconsistencies due to occlusions. Therefore, we re-warp the neighboring frames back to the reference frame, and utilize a threshold to filter out areas with significant errors. These areas, where the errors exceed the threshold, are considered as occluded regions:

\begin{equation}
\mathcal{L}_{geo} 
= \frac{1}{\mathbb{|V|}} \sum_{\bm{p}_r \in \mathbb{V}} || \tilde{\bm{p}}_r - \bm{H}_{nr} \bm{H}_{rn} \tilde{\bm{p}}_r ||.
\end{equation}

Finally, the multi-view consistent constrain consists of two components, 
the multi-view photometric constraint and the multi-view geometric consistency constraint:

\begin{equation}
\mathcal{L}_{mv} = \mathcal{L}_{ncc} + \mathcal{L}_{geo}.
\end{equation}
As shown in Figure~\ref{fig:ablation}, our method demonstrates that multi-view regularization is crucial for reconstruction accuracy. In summary, our overall set of constraints is as follows:

\begin{equation}
\mathcal{L} = \mathcal{L}_{flatten} + \mathcal{L}_{app} + \mathcal{L}_{local} + \mathcal{L}_{mv}.
\end{equation}

\subsection{Mesh Extraction}
\label{sec:meshextraction}

By incorporating our regularization term, we facilitate the generation of a mesh from the optimized Gaussian model. 
Subsequently, we proceed to render both a visual rendering and a depth map from various vantage points. 
These rendered images and depth maps are then utilized to fusion into a projected truncated signed distance function (TSDF) volume \cite{zeng20173dmatch} to finally create the superior quality 3D surface meshes and point clouds.

%% file: Sections/5-experiment.tex
\section{Experiments}
\label{sec:experiments}

\begin{figure*}[htbp]
    \begin{center}
        \centerline{\includegraphics[width=\linewidth]{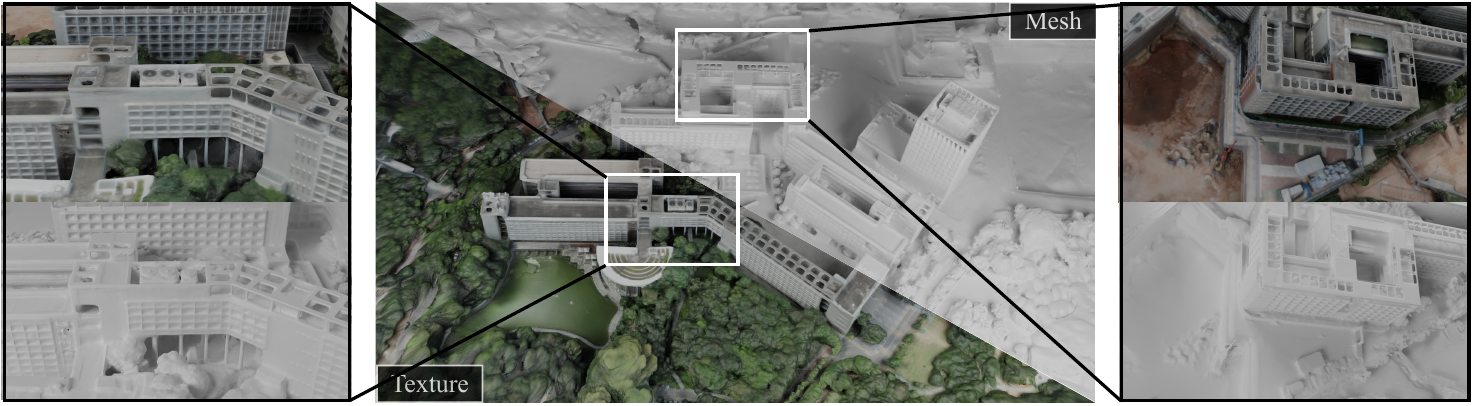}}
        \caption{\textbf{Visualization of surface reconstruction.} The figures showcase the results of training GigaGS on real aerial scenes, followed by rendering multiple RGB and depth maps, and ultimately obtaining the surface reconstruction results using TSDFusion \cite{zeng20173dmatch}.}
        \label{fig:mesh}
    \end{center}
\end{figure*}

\definecolor{pink}{RGB}{250,193,192}
\definecolor{yellow}{RGB}{254,249,207}

\paragraph{Implement Details} 
\label{sec:implement}

In our experiment, we reduced the side length of 4K aerial images to one-fourth of their original size and aligned them with a comparative method. 
Subsequently, we employed pixel-sfm \cite{lindenberger2021pixsfm} to obtain an initial point cloud from the aerial images and performed Manhattan world alignment, aligning the $y$-axis perpendicular to the world coordinate axis of the ground plane.
We divided the entire scene into $4 \times 2$ partitions in the case of rubble, building, residence, and sci-art, while for the largest scene, campus, we divided it into $4 \times 4$ partitions. Each partition was subjected to training for 120,000 iterations to ensure sufficient convergence.
Upon the completion of independent training for each partition, we discard all anchors except for those in the original partition $c$. 
This approach ensures that each partition is ultimately non-overlapping, thereby enabling the construction of a comprehensive scene.

\paragraph{Baseline Methods} 
\label{sec:baseline}

For the purpose of comparing our surface reconstruction results, we have selected Neuralangelo \cite{li2023neuralangelo}, NeuS \cite{wang2021neus}, PGSR \cite{chen2024pgsr}, and SuGaR \cite{guedon2023sugar} as the comparative methods. Neuralangelo and Neus are methods rooted in the Nerf framework, whereas Sugar and PGSR are methods that relies on 3DGS. Furthermore, to supplement the aforementioned methods, we have included VastGaussian \cite{lin2024vastgaussian} and MegaNeRF \cite{turki2022meganerf} as additional comparative methods in the analysis of rendering outcomes.

\begin{table*}[t]
\begin{center}
    \caption{ \textbf{Quantitative results of rendering quality.} 
    We report SSIM$\uparrow$, PSNR$\uparrow$ and LPIPS$\downarrow$ on test views.
    \colorbox{pink}{"Red"}, and \colorbox{yellow}{"Yellow"} denote the best and second-best results. 
    "-"  indicates that training cannot proceed due to out of memory.}
    \label{tab:compare}
    \setlength{\tabcolsep}{0.25mm}
        \begin{tabular}{l ccc ccc ccc ccc ccc}
            \toprule[1.1pt]
             &   \multicolumn{3}{c}{\emph{Building}}  
             &   \multicolumn{3}{c}{\emph{Rubble}} 
             &   \multicolumn{3}{c}{\emph{Campus}} 
             &  \multicolumn{3}{c}{\emph{Residence}} 
             &   \multicolumn{3}{c}{\emph{Sci-Art}} \\
             \cmidrule(r){2-4} \cmidrule(r){5-7} \cmidrule(r){8-10} \cmidrule(r){11-13} \cmidrule(r){14-16} 
            &  SSIM & PSNR & LPIPS   
            &  SSIM & PSNR & LPIPS 
            &  SSIM & PSNR & LPIPS   
            &  SSIM & PSNR & LPIPS   
            &  SSIM & PSNR & LPIPS  \\
            \midrule
            \textbf{No mesh (except GigaGS)}  \\
            \midrule
            Mega-NeRF \*   & \cellcolor{white}0.569 & \cellcolor{white}21.48 & \cellcolor{white}0.378 
                           & \cellcolor{white}0.575 & \cellcolor{white}24.70 & \cellcolor{white}0.407 
                           & \cellcolor{white}0.561 & \cellcolor{yellow}23.93 & \cellcolor{white}0.513 
                           & \cellcolor{white}0.648 & \cellcolor{yellow}22.86 & \cellcolor{white}0.330 
                           & \cellcolor{yellow}0.769 & \cellcolor{yellow}26.25 & \cellcolor{white}0.263 \\
                           
            VastGaussian \*  & \cellcolor{yellow}0.804 & \cellcolor{yellow}23.50 & \cellcolor{yellow}0.130 
                             & \cellcolor{yellow}0.823 & \cellcolor{pink}26.92 & \cellcolor{pink}0.132 
                             & \cellcolor{pink}0.816 & \cellcolor{pink}26.00 & \cellcolor{pink}0.151 
                             & \cellcolor{pink}0.852 & \cellcolor{pink}24.25 & \cellcolor{pink}0.124 
                             & \cellcolor{pink}0.885 & \cellcolor{pink}26.81 & \cellcolor{pink}0.121 \\

            GigaGS(ours) & \cellcolor{pink}0.905 & \cellcolor{pink}26.69 & \cellcolor{pink}0.125 
                         & \cellcolor{pink}0.837 & \cellcolor{yellow}25.10 & \cellcolor{yellow}0.167 
                         & \cellcolor{yellow}0.773 & \cellcolor{white}22.79 & \cellcolor{yellow}0.254
                         & \cellcolor{yellow}0.822 & \cellcolor{white}22.30 & \cellcolor{yellow}0.190
                         & \cellcolor{yellow}0.883 & \cellcolor{white}24.34 & \cellcolor{yellow}0.158 \\

            \midrule

            \textbf{With mesh}  \\
            \midrule
            
            PGSR         
                         & \cellcolor{white}0.480 & 16.12 & 0.573
                         & \cellcolor{yellow}0.728 & \cellcolor{yellow}23.09 & \cellcolor{white}0.334
                         & 0.399 & 14.02 & \cellcolor{white}0.721
                         & \cellcolor{yellow}0.746 & \cellcolor{yellow}20.57 & \cellcolor{white}0.289
                         & \cellcolor{yellow}0.799 & \cellcolor{yellow}19.72 & \cellcolor{white}0.275 \\

            SuGaR        & 0.507 & 17.76 & \cellcolor{white}0.455
                         & \cellcolor{white}0.577 & \cellcolor{white}20.69 & 0.453
                         & - & - & -
                         & 0.603 & \cellcolor{white}18.74 & 0.406
                         & 0.698 & 18.60 & 0.349 \\

            NeuS         & 0.463 & \cellcolor{yellow}18.01 & 0.611
                         & 0.480 & 20.46 & 0.618
                         & 0.412 & \cellcolor{white}14.84 & 0.709
                         & 0.503 & 17.85 & 0.533
                         & 0.633 & 18.62 & 0.472 \\

            Neuralangelo & \cellcolor{yellow}0.582 & \cellcolor{white}17.89 & \cellcolor{yellow}0.322
                         & 0.625 & 20.18 & \cellcolor{yellow}0.314
                         & \cellcolor{yellow}0.607 & \cellcolor{yellow}19.48 & \cellcolor{yellow}0.373
                         & \cellcolor{white}0.644 & 18.03 & \cellcolor{yellow}0.263
                         & \cellcolor{white}0.769 & \cellcolor{white}19.10 & \cellcolor{yellow}0.231 \\
                         
            GigaGS(ours) & \cellcolor{pink}0.905 & \cellcolor{pink}26.69 & \cellcolor{pink}0.125 
                         & \cellcolor{pink}0.837 & \cellcolor{pink}25.10 & \cellcolor{pink}0.167 
                         & \cellcolor{pink}0.773 & \cellcolor{pink}22.79 & \cellcolor{pink}0.254
                         & \cellcolor{pink}0.822 & \cellcolor{pink}22.30 & \cellcolor{pink}0.190
                         & \cellcolor{pink}0.883 & \cellcolor{pink}24.34 & \cellcolor{pink}0.158 \\

            \bottomrule[1.1pt]
        \end{tabular}
\end{center}
\centering
\end{table*}

\paragraph{Datasets and Metrics} 
\label{sec:datasets}

We employ GigaGS on datasets consisting of real-life aerial large-scale scenes, which encompass the \textit{Building} and \textit{Rubble} scenes extracted from Mill-19 \cite{turki2022meganerf}, along with the \textit{Sci-Art}, \textit{Campus}, and \textit{Residence} scenes sourced from Urbanscene3d \cite{urbanscene3d}.
To maintain consistency, we employ the same dataset partitioning as MegaNeRF \cite{turki2022meganerf}.
We utilized visual quality metrics, namely PSNR, SSIM, and LPIPS \cite{zhang2018unreasonable}, to compare the rendering quality on the test set. 
Additionally, we compared the visualizations of the results on the test set with methods capable of extracting depth and normal maps.


\subsection{Visual Quality}
\label{sec:visual_quality}

The figure \ref{fig:visual} compares the rendered results of the novel perspective reconstruction method along with their corresponding depth maps and normal maps. 
It can be observed that our GigaGS outperforms existing surface reconstruction methods in terms of surface textures and scene geometry. 
The existing methods based on NeRF \cite{li2023neuralangelo, wang2021neus} lack fine details and exhibit blurry and erroneous structures in image rendering. 
Similarly, the existing methods based on 3DGS \cite{chen2024pgsr, guedon2023sugar} are plagued by artifacts, resulting in undesirable rendering outcomes.
In the table \ref{tab:compare}, we quantitatively compare the test set results of the aforementioned methods, along with the inclusion of two additional large-scale novel view synthesis (NVS) methods, solely for the purpose of comparing rendering quality. 
It can be observed that our GigaGS method significantly improves the rendering quantitative results of existing surface reconstruction methods, while achieving comparable performance to the NVS method.

\subsection{Mesh Reconstruction}
\label{sec:meshreconstruction}

We utilized the method mentioned in the section \ref{sec:meshextraction} to extract a mesh from GigaGS. 
As shown in the figure \ref{fig:mesh}, our approach enables the extraction of high-quality meshes while ensuring high-quality rendering. 
This capability holds potential to support a wide range of applications, such as navigation, simulation, and virtual reality (VR).

\subsection{Quantity of 3D Gaussian Splatting}
\label{sec:quantityof3dgs}

\begin{figure}[htbp]
    \begin{center}
        \centerline{\includegraphics[width=\linewidth]{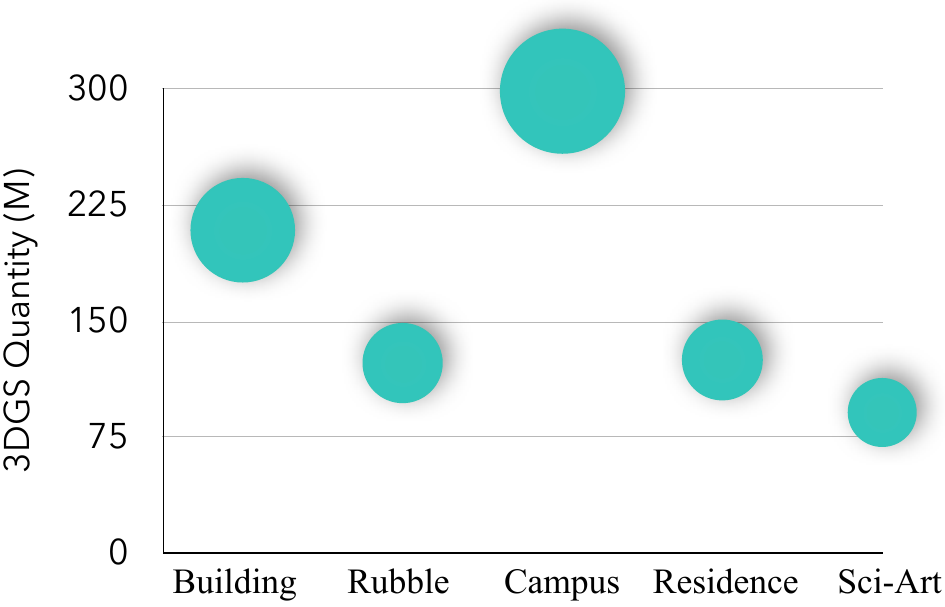}}
        \caption{\textbf{Quantity of 3D Gaussian Splatting in five scenes.} This figure shows the actual number of 3DGS obtained by Our method.}
        \label{fig:number}
    \end{center}
    \vspace{-1cm}
\end{figure}

In the figure \ref{fig:number}, we illustrate the quantity of 3DGS obtained through optimization in various scenarios. Owing to our stratified scene representation and partition-based optimization strategy, we are able to represent scenes with an increased number of 3DGS. As the volume of scene data in practical applications grows, it can even approach the giga-level. Consequently, GigaGS can maximize the capture of scene details.

%% file: Sections/6-conclusion.tex
\section{Conclusion}
\label{sec:conclusion}
In this paper, we propose \NickName. To the best of our knowledge, \NickName \space is the first work for large-scale scene surface reconstruction with 3D Gaussian Splatting. Through careful design, \NickName \space deliver high quality 3D surface and can process large scenes in parallel.

\paragraph{Limitation} The performance of 3D Gaussian is highly correlated to the performance of COLMAP, which may degrade the performance especially for textureless regions.